\documentclass[lettersize,journal]{IEEEtran}
\usepackage{amsmath,amsfonts}
\usepackage{algorithmic}
\usepackage{algorithm}
\usepackage{array}
\usepackage{textcomp}
\usepackage{stfloats}
\usepackage{url}
\usepackage{verbatim}
\usepackage{graphicx}
\usepackage{cite}
\usepackage{xcolor}
\usepackage{caption}
\usepackage[hidelinks]{hyperref}
\usepackage{subcaption}
\usepackage{wrapfig}
\usepackage{booktabs}
\usepackage{enumitem}
\usepackage{graphicx}
\usepackage{tabularx}
\usepackage{makecell}
\usepackage{subcaption}
\usepackage{caption}
\usepackage{bbm}
\usepackage{titlesec}

\titlespacing{\subsection}{0pt}{\parskip}{-\parskip}
\begin{document}



\addtolength{\textfloatsep}{-0.25in}
\title{Towards Model-Size Agnostic, Compute-Free, Memorization-based Inference of Deep Learning}
\author{Davide Giacomini$^1$, Maeesha Binte Hashem$^1$, Jeremiah Suarez$^2$, Swarup Bhunia$^3$,
and Amit Ranjan Trivedi$^1$ \\
$^1$AEON Lab, University of Illinois at Chicago (UIC), Chicago, IL, USA, $^2$Illinois Mathematics and Science Academy, IL, USA, $^3$University of Florida (UFL), FL, USA}
\maketitle

\begin{abstract}
The rapid advancement of deep neural networks has significantly improved various tasks, such as image and speech recognition. However, as the complexity of these models increases, so does the computational cost and the number of parameters, making it difficult to deploy them on resource-constrained devices. This paper proposes a novel memorization-based inference (MBI) that is \textit{compute-free} and only requires lookups. Specifically, our work capitalizes on the inference mechanism of the recurrent attention model (RAM), where only a small window of input domain (glimpse) is processed in a one-time step, and the outputs from multiple glimpses are combined through a hidden vector to determine the overall classification output of the problem. By leveraging the low-dimensionality of glimpse, our inference procedure stores key-value pairs comprising of glimpse location, patch vector, \textit{etc.} in a table. The computations are obviated during inference by utilizing the table to read out key-value pairs and performing compute-free inference by memorization. By exploiting Bayesian optimization and clustering, the necessary lookups are reduced, and accuracy is improved. We also present in-memory computing circuits to quickly look up the matching key vector to an input query. Compared to competitive compute-in-memory (CIM) approaches, MBI improves energy efficiency by $\sim$2.7$\times$ than multilayer perceptions (MLP)-CIM and by $\sim$83$\times$ than ResNet20-CIM for MNIST character recognition.  
\end{abstract}


\section{Introduction}
Ultra-low-power edge inference of deep neural networks (DNNs) has revolutionized many application spaces, enabling edge devices to perform complex data-driven inference and real-time decision-making with minimal energy consumption. The edge inference of DNNs has opened up new avenues for applications such as wearables, smart homes, Internet-of-Things (IoT), cyber-physical systems, and many more\cite{8763885}. By performing most computations at the data source, edge inference also helps mitigate privacy and security concerns by keeping sensitive information on local devices rather than transmitting it to remote servers. Additionally, edge computing helps to reduce network congestion and lowers carbon footprint by minimizing the need for data to be transmitted over long distances. 

\begin{figure}[t]
    \centering  
    \includegraphics[width=\linewidth]{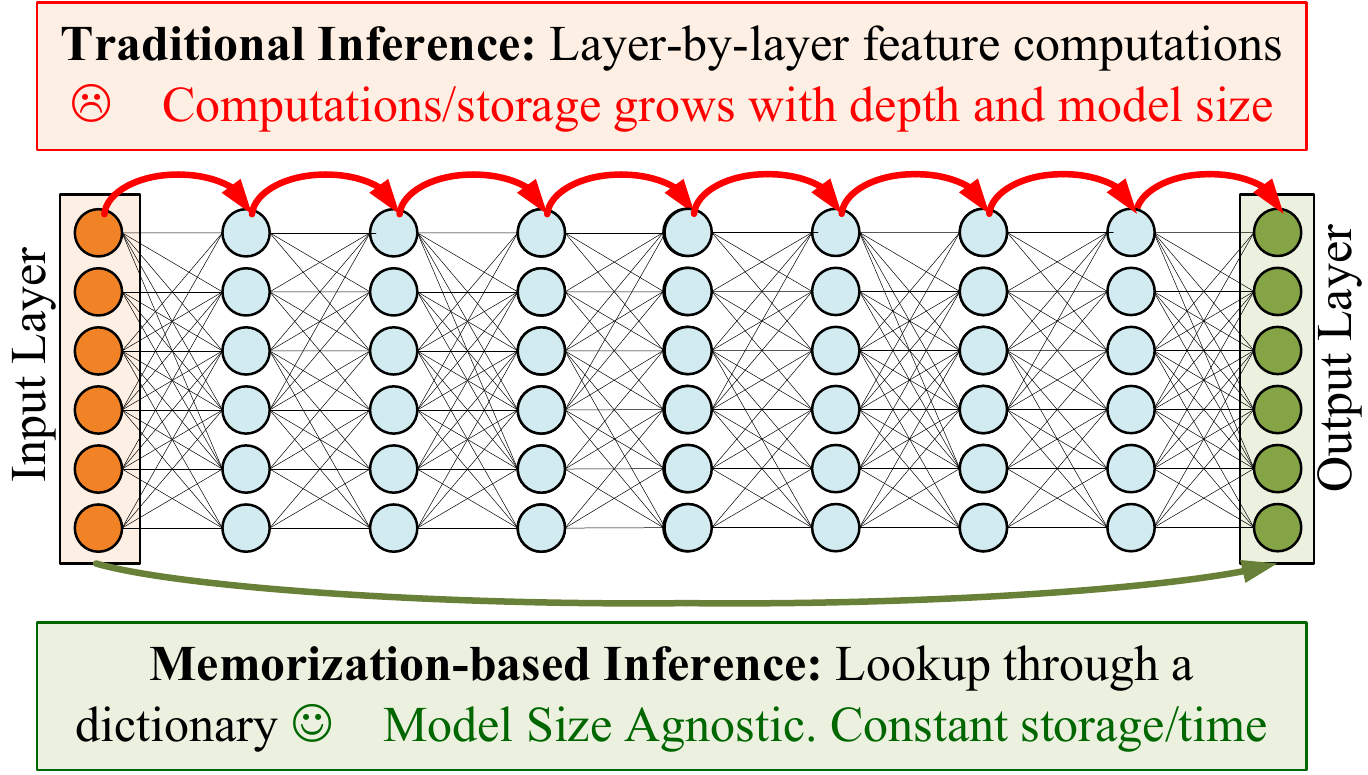}
    \captionsetup{justification=justified}
    \captionsetup{skip=2pt}
    \label{fig:figure1}
    \caption{\small \textbf{Memorization-based Inference (MBI) \textit{vs.} Traditional Inference:} We propose a novel memorization-based inference (MBI) that is \textit{compute-free} and only requires lookups for inference. While, under traditional inference, storage/computing costs increase proportionally to the number of layers and depth of DNN, MBI is agnostic to the model size and complexity. }    
\end{figure}


DNNs are increasingly utilized in applications like autonomous insect-scale drones\cite{shukla2021ultralow}, robotic surgery \cite{shukla2022mc}, and cognitive assistants. However, improving their predictive capacity in complex signal spaces requires increasing the number of trainable parameters and network depth. For instance, GPT models have achieved remarkable performance but require enormous parameters, ranging from 175 billion in GPT-3 to 100 trillion in GPT-4. Edge-friendly models like MobileNetV2 \cite{sandler2018mobilenetv2}, ResNet50 \cite{zagoruyko2016wide}, and EfficientNet-B0 \cite{tan2019efficientnet} offer some efficiency, but still demand significant computational resources. Limited resources on edge devices pose challenges in handling the growing complexity of deep learning models.

\begin{figure*}[t]
    \centering  
    \includegraphics[width=\linewidth]{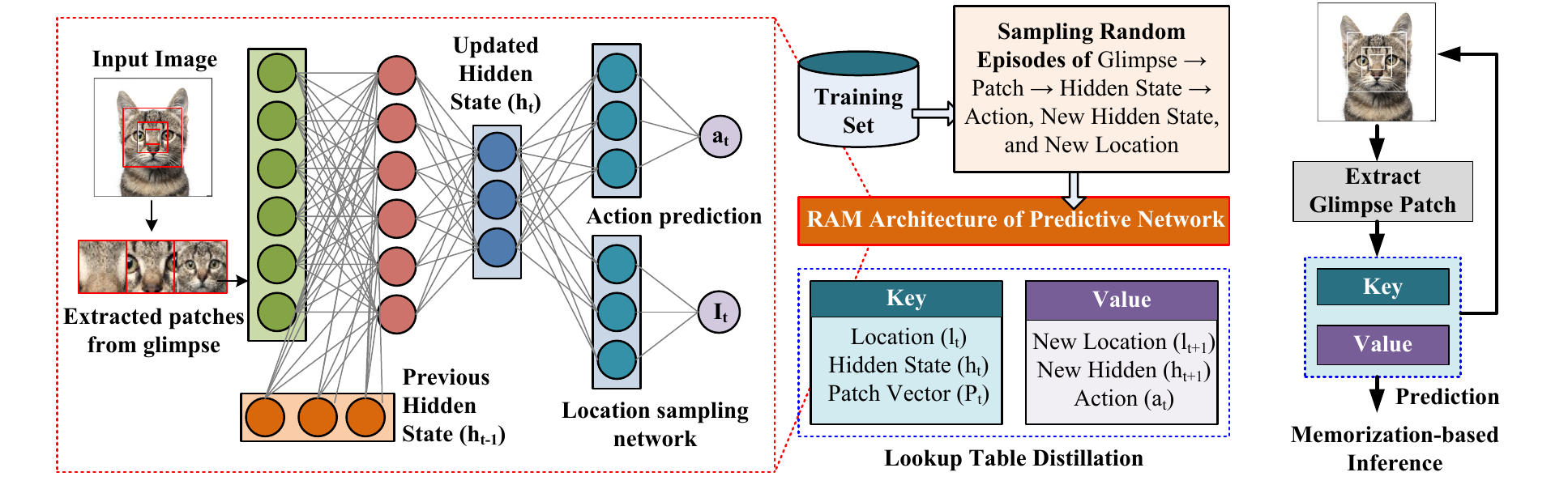}
    \captionsetup{justification=justified}
    \captionsetup{skip=2pt}
    \captionsetup{belowskip=-5pt}
    \caption{\small \textbf{Overview of memorization-based inference (MBI):} To minimize the size of LUTs for memorization, we capitalize on recurrent attention model (RAM) architecture of neural network inference where only a low-dimensional glimpse of input is processed in one time-step. Our approach distills LUT from RAM architecture. During inference, only a sequence of key-value readouts from LUTs is necessary. }
    \label{fig:figure1}
\end{figure*}

Fundamentally, there could be two ways to perform inferential computations [Fig. 1]. The first approach involves processing all necessary arithmetics through modules, such as multipliers, adders, shifters, and other components, to obtain the resultant output. The second approach involves \textit{memorization} where the resultant output is precomputed and memorized at all possible input combinations and thereafter retrieved during inference, obviating the need for any computations. Notably, the second approach becomes increasingly attractive as the workload of inferential computations increases. With the phenomenal growth of DNN model sizes and the number of model parameters reaching billions and trillions, the second approach might also be more memory-efficient by only storing a lookup table (LUT) of input-output (key-value) combinations than the DNN model parameters themselves.

In this work, we pay closer attention to the above model-size agnostic \textit{memorization-based inference}, i.e., MBI of DNNs to explore pathways for disruptive enhancement of edge inference. Specifically, our work makes the following contributions: 
\begin{itemize}[noitemsep,topsep=0pt,leftmargin=4pt]
\item We introduce a novel \textit{memorization-based inference} (MBI), which involves distilling a pre-trained model into a LUT to perform inference without requiring intensive arithmetics such as multiplications or additions. Instead, the inference process is \textit{compute-free}, relying only on a sequence of key-value lookups on the distilled LUT for a given input query. 
\item To improve the scalability of MBI, we demonstrate a novel framework combining recurrent attention mechanisms, Bayesian optimization-based optimal distance metric search, hierarchical clustering, and in-memory determination of the closest entry to an input. Recurrent attention mechanisms are leveraged to minimize the size of LUTs. Bayesian optimization of distance metrics improves prediction accuracy with incomplete tables. Hierarchical clustering minimizes the table size for each lookup. Finally, in-memory determination of the closest key to the input query improves the speed.
\item We characterize MBI for character recognition on the MNIST dataset under extremely low precision. The hidden state vector is quantized to one bit and the input patch vector to two bits. Specifically, we demonstrated a \textit{mixed-memorization-based inference} where most low-complexity images are processed through memorization and fewer high-complexity images require full processing of traditional machine learning. Compared to competitive compute-in-memory (CIM) approaches, MBI improves energy efficiency by $\sim$2.7$\times$ than multilayer perceptions (MLP)-CIM and by $\sim$83$\times$ than ResNet20-CIM for MNIST character recognition. 
\end{itemize}

Sec. II introduces the opportunities and challenges for MBI. Sec. III details various components of the inference methodology and presents simulation results. Sec. IV concludes.  

\section{Model-Size Agnostic, Compute-Free, Memorization-based Inference (MBI)}

\subsection{Opportunities and Challenges of Lookup-Only Inference}
Our approach is focused on developing an inference methodology that can make predictive workloads independent of the number of layers and model parameters. This would allow for complex predictions to be made within a \textit{constant time and memory budget}. In Fig. 2, our MBI approach accomplishes this by distilling the model's predictions on query inputs into a key-value LUT, which requires only searching a matching key to query to make predictions, thereby avoiding intermediate feature extractions.

Despite the potential for constant time/storage predictions, independent of the predictive model's architecture and parameters, naive memorization of even simpler prediction tasks results in extremely large LUT sizes that cannot be practically synthesized, stored, or inferred. For instance, consider character recognition on the MNIST dataset, where each image is 28$\times$28 pixels \cite{lecun1998gradient}. Even if we consider a 2-bit representation of each pixel value in character images, a binarized vector representing the input image would be 2$\times$28$\times$28 = 1568 bits long, requiring a complete table with 1.04e+333 number of rows for all possible inputs! As the bit precision of input images (such as 8 bits per pixel) or the size of input images (such as 224$\times$224 for cropped ImageNet images) increases, the size of memorization tables becomes even more exorbitant. Thus, although the potential benefits of constant time and memory predictions are evident under MBI, naive memorization is infeasible even on simpler predictive tasks.

\begin{figure*}[t]
  \centering
  \begin{subfigure}{0.32\textwidth}
    \centering
    \includegraphics[width=\textwidth]{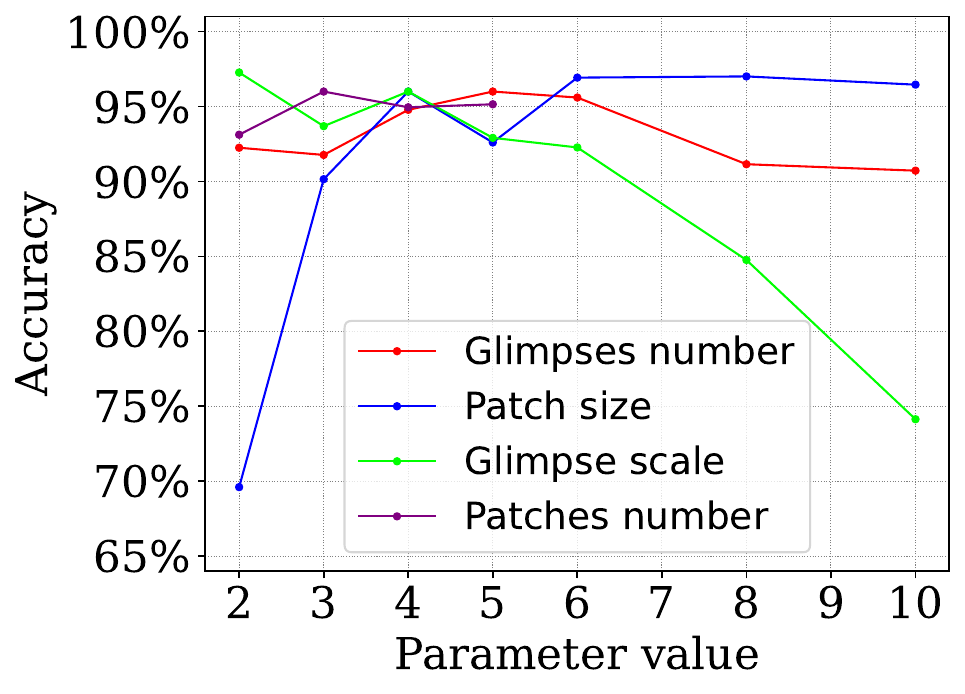}
    \caption{}
    \label{fig:patch}
  \end{subfigure}
  \begin{subfigure}{0.32\textwidth}
    \centering
    \includegraphics[width=\textwidth]{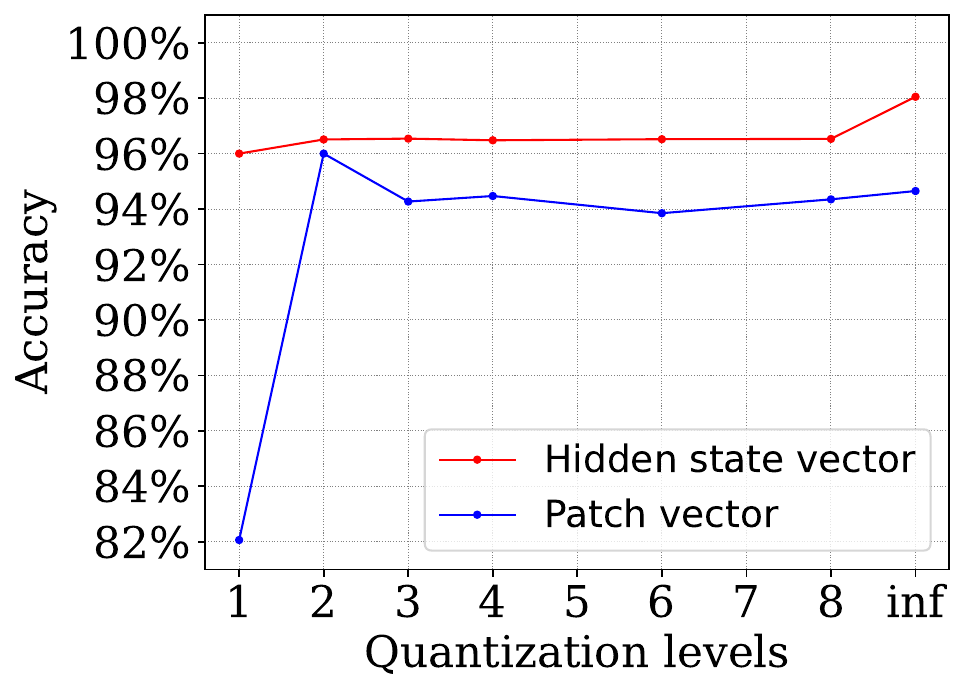}
    \caption{}
    \label{fig:quantization}
  \end{subfigure}
  \begin{subfigure}{0.32\textwidth}
    \centering
    \includegraphics[width=\textwidth]{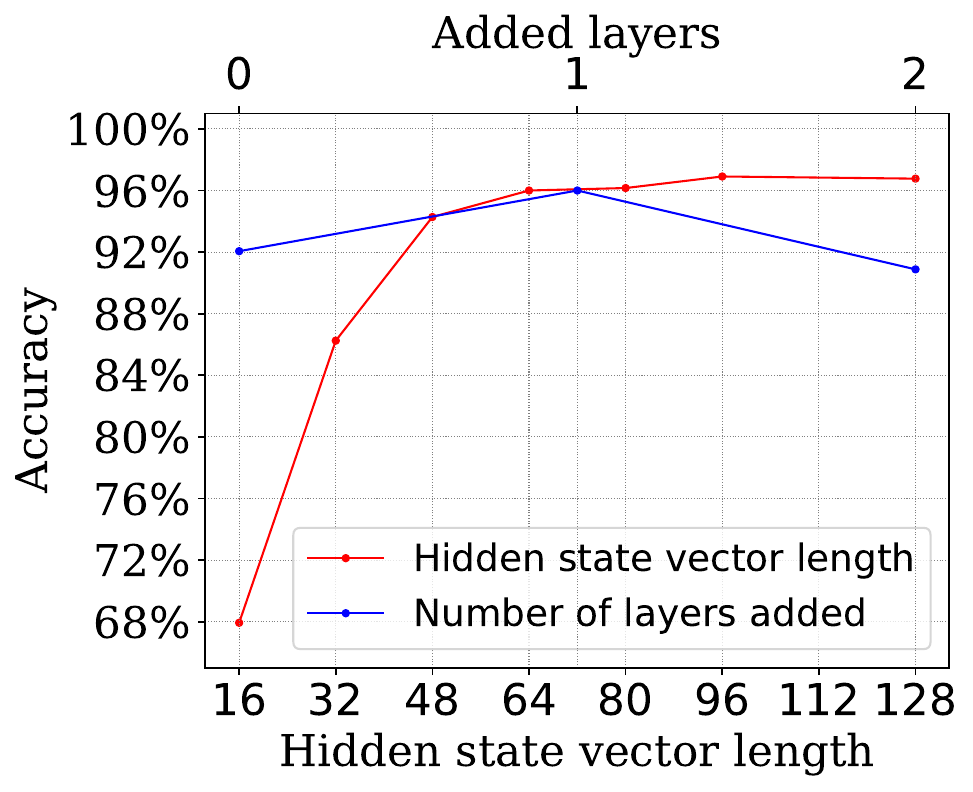}
    \caption{}
    \label{fig:structure}
  \end{subfigure}
  \captionsetup{justification=justified}
  \captionsetup{belowskip=-10pt}
  \caption{\small \textbf{Hyperparameter and quantization space explorations of RAM architecture for MBI:} For character recognition on MNIST, \textbf{(a)} accuracy at the varying number of glimpses, patch size, the scale of glimpse, and the number of patches utilized in each time-step. \textbf{(b)} Accuracy at varying hidden and patch vector quantization levels. \textbf{(c)} Accuracy at the varying length of the hidden vector and the increasing number of additional layers for hidden state determination.}
  \label{fig:hyperparameters}
\end{figure*}

\subsection{Proposed Methodology for Memorization-based Inference}
Our methodology employs several techniques to enhance the feasibility of constant time/storage-bound MBI. Fig. 2 presents the overview of the proposed methodology. Details on various components will be presented in the subsequent section. \textit{Primarily,} we leverage the recurrent attention model (RAM) architecture of neural networks, introduced in \cite{mnih2014recurrent}, where a recurrent neural network is integrated with attention mechanisms. The attention mechanism of RAM allows the network to assign different weights to different parts of the input so that it can selectively attend only to the most salient information of the input. The RAM is designed to learn to focus its attention on only a low-dimensional glimpse of the input image. In Fig. 2, for MBI, the RAM architecture minimizes the operating input dimension in each time-step (such as to only 3$\times$3), thus enabling a significant reduction in the necessary LUT to make the inference scheme feasible.

\textit{Secondly,} we rely on \textit{incomplete tables} for MBI where only the closest match, instead of an exact match, to an input query vector is required to determine the readout. The LUT size for MBI need not span the entire input space; instead, it can just be a subset of the input space. For example, under the LUT size budget of $N$ rows, the input space can be sampled on $N$ query points, and an incomplete table of $N$ rows can be used for MBI. Enabling MBI from incomplete tables further improves the practicality of the procedure where storage resources can be explicitly accounted for. Furthermore, we explore Bayesian optimization to determine the optimal distance metrics to improve inference accuracy even with incomplete tables. Bayesian optimization can optimize expensive-to-evaluate functions, such as optimal distance metric in MBI, by building a probabilistic objective function model and iteratively selecting new points to evaluate based on the expected improvement in the model's performance. Our results in the next section indicate that Bayesian optimization-based optimal distance metrics can significantly improve the prediction accuracy by 3-4\%.      

\textit{Finally}, 
to enhance MBI's speed and energy efficiency, we utilize hierarchical clustering and analog-domain in-memory determination with flexible distance metrics. Hierarchical K-means clustering organizes table entries into a search tree, enabling quick search of a subset of the table at the leaf node for the closest entry to a query. Analog-domain in-memory computing circuits compute distances in parallel, using a winner-takes-all (WTA) approach for rapid retrieval. Performing computations within the memory array eliminates data movements, reducing energy and latency overheads. We analyze the impact of non-idealities, like transistor process variation, on inference accuracy in the analog circuit components at the leaf of the clustering tree.
\section{Components of Memorization-based Inference Methodology and Simulation Results}

\subsection{Recurrent Attention Mechanisms for Downscaling LUTs}
Fig. 2 provides an overview of our RAM architecture for memory-based inference. The attention mechanism in the figure utilizes a glimpse network to extract a smaller window of the input image for further processing. Recurrence is achieved through a core network that processes the hidden state vector from the previous time step and outputs a new hidden state vector for the current time step. 

The location network computes a new location vector at each time step, enabling focus on specific image regions. High-resolution patches are extracted from the selected locations, progressively increasing in size at lower resolutions to widen the network's image coverage. These patches are stacked to form a patch vector, processed further through a linear layer. The glimpse network combines the glimpse vector with the previous time step's hidden state and feeds it into the core network. The core network output is propagated to the next time step and to the location and classification networks.

The size of LUTs and the number of lookups in our MBI are influenced by RAM architecture parameters: glimpses, patch size, glimpse scale, and the number of patches. More glimpses lead to increased hidden state updates and, consequently, more MBI lookups. Larger patch sizes and more patches require longer key vectors, leading to larger LUTs and more rows for comprehensive key coverage. Increasing glimpse scale allows global image feature awareness but sacrifices granularity by compressing to a low-resolution window.

\begin{figure*}[t]
  \centering
  \begin{subfigure}{0.30\textwidth}
    \centering
    \includegraphics[width=\textwidth]{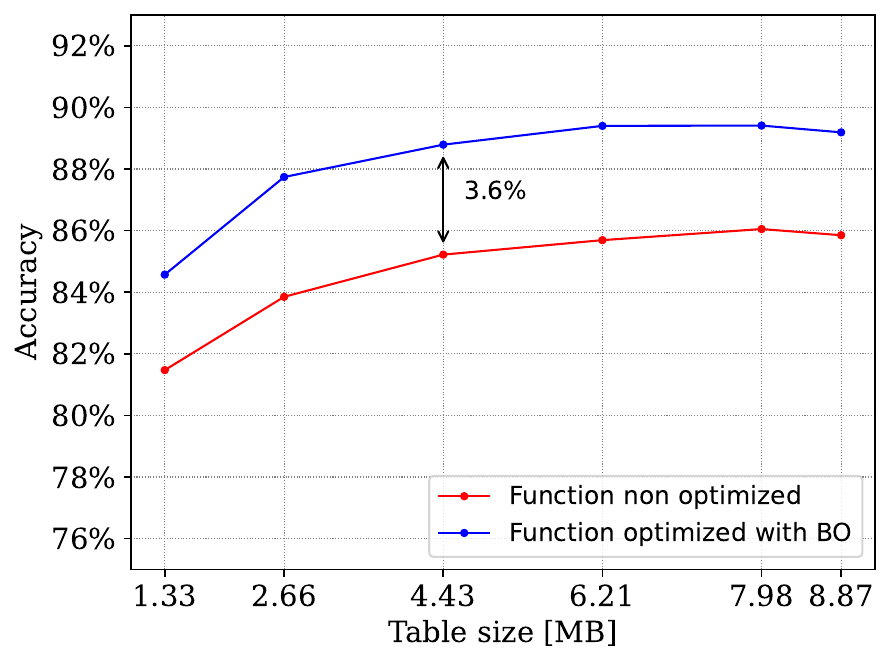}
    \caption{}
    \label{fig:patch}
  \end{subfigure}
  \begin{subfigure}{0.33\textwidth}
    \centering
    \includegraphics[width=\textwidth]{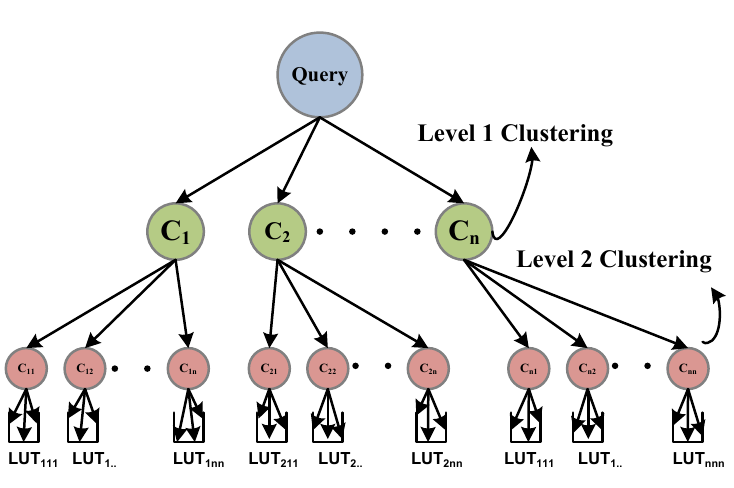}
    \caption{}
    \label{fig:quantization}
  \end{subfigure}
  \begin{subfigure}{0.32\textwidth}
    \centering
    \includegraphics[width=\textwidth]{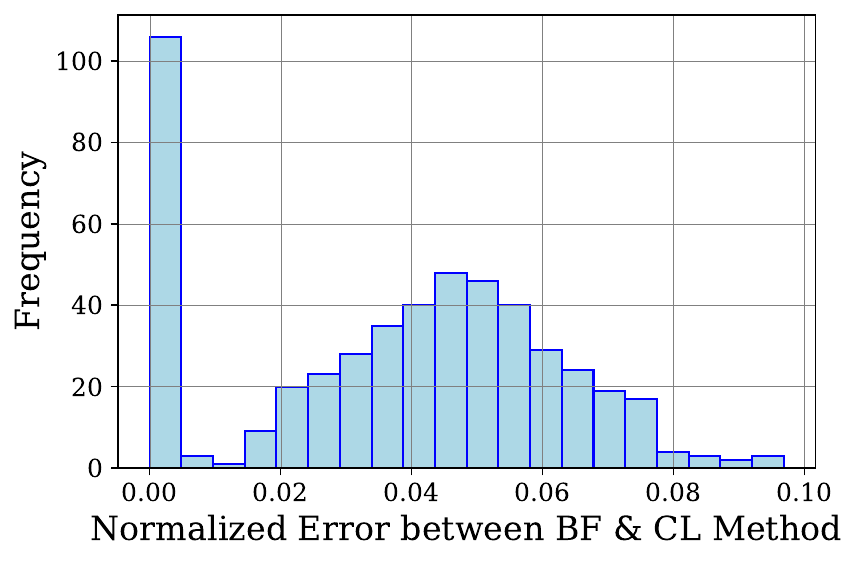}
    \caption{}
    \label{fig:structure}
  \end{subfigure}
  \captionsetup{justification=justified}
  \captionsetup{belowskip=-10pt}
  \caption{\small \textbf{Improving memory-efficiency of MBI:} \textbf{(a)} Under MBI with incomplete tables, the prediction accuracy for MNIST at varying LUT size under Bayesian optimization (BO)-based weighted distance \textit{vs.} unweighted metrics. BO improves prediction accuracy by 3--4\% across LUT size. \textbf{(b)} Hierarchical clustering to minimize necessary comparisons for finding the matching key to a query. \textbf{(c)}. Histogram of key matching error between brute force (BF) and clustering (CL)-based matching key search. }
  \label{fig:hyperparameters}
\end{figure*}


To develop the MBI table, we fine-tuned hyperparameters and quantization levels to enhance memory efficiency. Fig. 3(a) demonstrates the impact of different hyperparameters on accuracy. Increasing patch size initially improves accuracy but saturates beyond a certain point. Glimpse scale increase reduces accuracy as it tries to capture the entire image in a few pixels. Similarly, accuracy saturation occurs with more glimpses and patches. Fig. 3(c) shows that adding one extra layer to the original network achieves the highest accuracy, while further layers decrease accuracy. Additionally, larger hidden state sizes improve accuracy by capturing more information.
\begin{figure}
    \centering
    \includegraphics[width=0.5\textwidth]{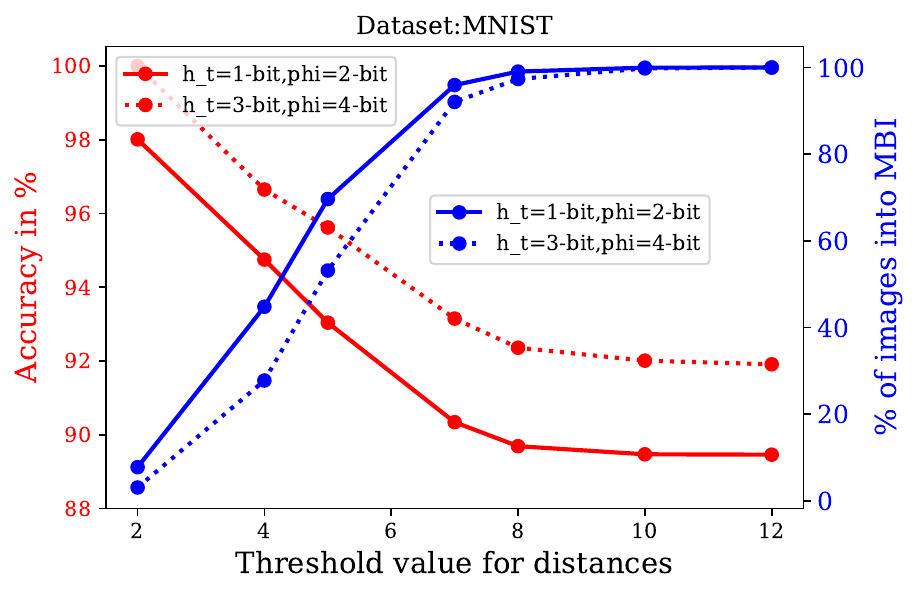}
    \caption{Mixed-MBI at sweeping mixing threshold.}
    \captionsetup{justification=justified}
    \captionsetup{belowskip=-15pt}
    \label{fig:example}
\end{figure}

In Fig. 3(b), the accuracy remains relatively constant across the range of increasing patch size quantization bit sizes, except for a single-bit quantization. This deviation in accuracy can be attributed to a significant loss of information at the input level of the model due to 1-bit quantization. In contrast, the accuracy remains stable concerning the increase in bit size for hidden state vector quantization.

\subsection{Bayesian Optimization for Optimal Distance Metric}
To optimize the workload of MBI, we employ incomplete tables, where only a limited number of input combinations are stored based on available storage resources. Through randomly sampling the input space and capturing input-output episodes from the main model, we distill this information onto the MBI table. Consequently, instead of exact matches, we retrieve information from incomplete tables by searching for the closest match to an input query.

\begin{figure*}[t]
    \centering  
    \includegraphics[width=\linewidth]{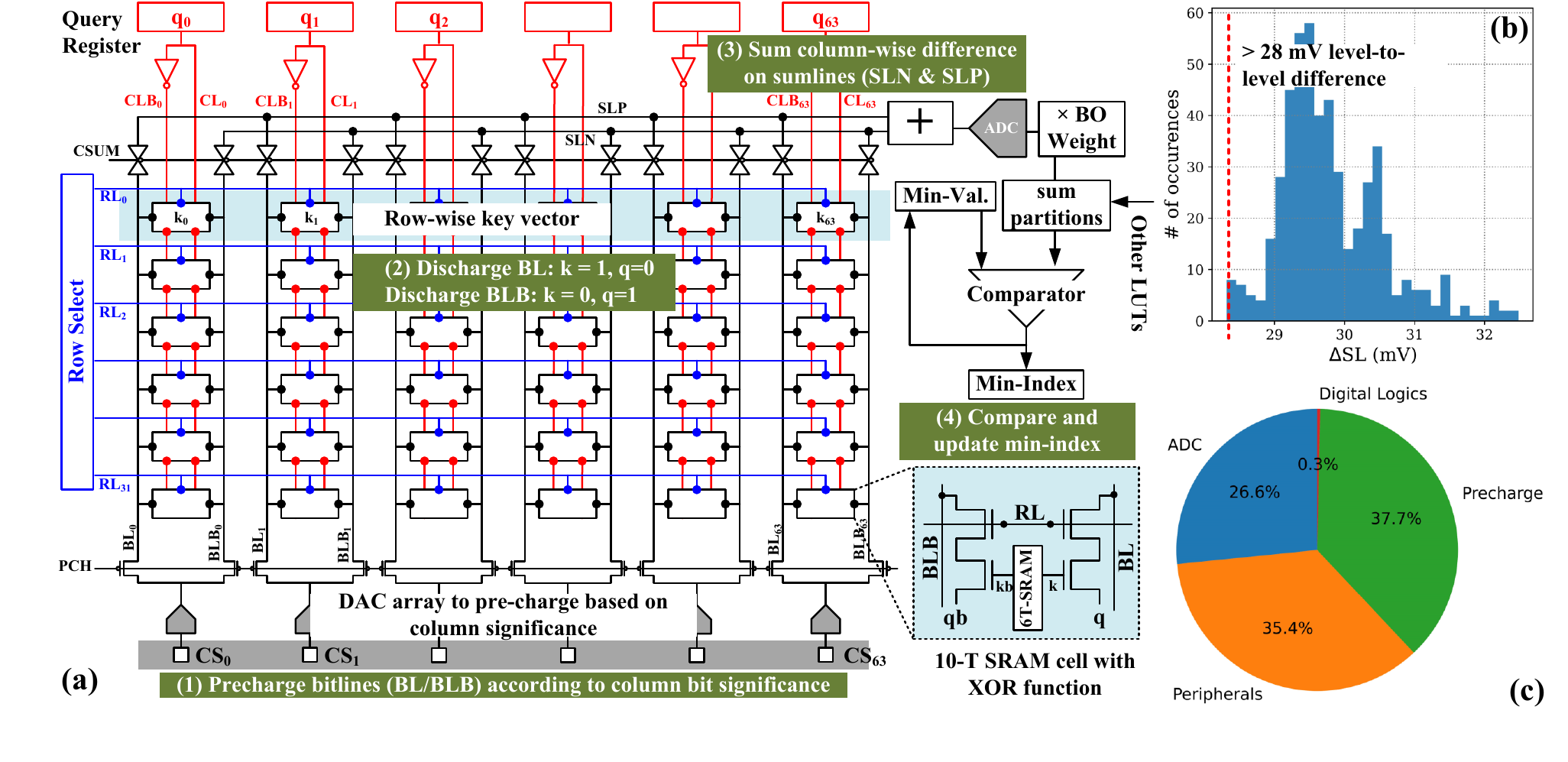}
    \captionsetup{justification=justified}
    \captionsetup{belowskip=-10pt}
    \caption{\small \textbf{In-memory search of matching key vector to the query:} \textbf{(a)} Compute-in-memory array to determine Manhattan distance of applied query vector at the top and row-wise stored key vectors in the memory array. The array utilizes ten transistor cells comprising 6-T SRAM and a distance computing port shown at the bottom right. Computed distance is digitized, multiplied to Bayesian learned weights, and then partitions of key/query vectors processed across multiple LUTs are combined. Key vector rows are sequentially scanned to find the minimum distance key vector. \textbf{(b)} The potential difference at the sum line (SL) between nearest distance levels under process-induced variability (minimum-sized NMOS and PMOS are simulated with $\sigma_{VTH}$ = 60 mV). \textbf{(c)} The power distribution among ADC, row activation peripherals, column pre-charging, and logic operations.}
    \label{fig:bo}
\end{figure*}

To search for the optimal distance metric, we employed Bayesian optimization. A parameterized distance metric function between query $\mathcal{Q}$ and key $\mathcal{K}$, $\mathcal{D}(\mathcal{Q},\mathcal{K})$, based on weighted Manhattan distance was used as following:
\begin{equation}
    \mathcal{D}(\mathcal{Q},\mathcal{K}) = \frac{a \cdot \mathcal{M}(q_p,k_p) + b \cdot \mathcal{M}(q_h,k_h) + c \cdot \mathcal{M}(q_l,k_l)}{a+b+c}
\end{equation}
Here, $a$, $b$, and $c$ are learnable weighting parameters for patch vector ($p$), hidden state ($h$), and location vector ($l$) component of the key. $\mathcal{M}()$ is the Manhattan distance function. $q_p$, $q_h$, and $q_l$ are patch, hidden, and location vector components of the query $\mathcal{Q}$. With similar subscript notations, the components of key $\mathcal{K}$ are defined. In Fig.~\ref{fig:bo}, Bayesian optimization-based optimally-weighted Manhattan distance metric improves the prediction accuracy by 3--4\% across LUT sizes compared to unweighted distance.

\subsection{Synthesizing Lookup Tree by Hierarchical Clustering}
Fig. 4(b) depicts a hierarchical K-means clustering method used to organize key values in a tree structure. The objective is to minimize search load when finding the closest match to an input query in the LUT. Keys in the table are divided into clusters, and their centroids are determined. Sub-clusters are then formed through multi-level clustering, continuing until a threshold for each node's total number of elements is exceeded. The query is compared to cluster centroids during the query-matching process, directing it to the appropriate branch. This comparison process repeats with sub-cluster centroids until the query reaches a leaf node, which is exhaustively compared to a few key vectors to find the closest match. Fig. 4(c) shows the histogram of the distance normalized to one between the matching key vector to a random query using the hierarchical clustering-based approach compared to an exhaustive search throughout the table. The former has a significantly smaller workload. As can be seen, the hierarchical clustering-based approach finds a matching key vector comparable to the exhaustive search with a very high probability -- the maximum distance between the matching keys in both cases is less than 10\%. The MBI approach can also be integrated with traditional DNNs to improve accuracy. In Fig. 5, MBI is only applied when a matching key to the input falls within a distance threshold for each glimpse iteration. Otherwise, traditional DNN is employed on such harder-to-generalize inputs. In the figure, as the matching distance threshold increases, more data can be processed using MBI, however, at the cost of lower accuracy. 

\subsection{In-Memory Search of Closest Key to Query}
We present an in-memory processing approach to efficiently search the closest key vector to an input query in Fig. 6(a). Integrating the search and storage in the same memory structure obviates the table-data movements to enhance efficiency.  

The operational sequence of the circuits is as follows: Multibit key vectors are stored row-wise, and query vectors are applied on the top ports in the figure. In \underline{step-1}, the pre-charge (PCH) mechanism is activated to charge all bit lines (BL/BLB) according to the column's bit-significance factor. For example, considering the $p$-bit precision of the query and key vector, a column operating on the bit significance factor $j$ $\in [0, p-1]$ is precharged with $V_{P,max}/2^{p-j-1}$. $V_{P,max}$ is the maximum precharge level. In Fig. 3(b), only a 2-bit precision of the patch vector provided sufficient accuracy; hence, we encode the patch component of a key vector with 2-bit precision. From Fig. 3(b), the hidden state component of the key vector is quantized to 1-bit for the discussed results. Therefore, in the implemented scheme to compute the distance of a 2-bit quantized patch component of query and key vector, the columns storing higher significance bits are precharged to $V_{P,max}$ and the columns storing the least significant bits are precharged to $V_{P,max}/2$. 

In \underline{step-2}, after selecting the required row of key-bit vectors, each memory cell computes the bitwise difference between the corresponding key bit ($k$) and the applied query bit ($q$): BL discharges only when $k$=1 and $q$=0; BLB discharges only when $k$=0 and $q$=1. Such column discharges are utilized for multi-bit Manhattan distance computations. For example, consider the Manhattan distance computation between $n$-element long key $\mathcal{K}$ and query $\mathcal{Q}$. Under $p$-bit precision, $\mathcal{K}$/$\mathcal{Q}$ is processed on $p \times n$ columns. The Manhattan distance between the i$^{th}$ element of $\mathcal{K}$ and $\mathcal{Q}$ is given by $|\mathcal{K}_i - \mathcal{Q}_i| = \sum_{j=0}^{p-1}|k_{ij} - q_{ij}| \times 2^{j}$. Here, $k_{ij}$ \& $q_{ij}$ are the corresponding binary bits of $\mathcal{K}$ and $\mathcal{Q}$. Therefore, if $k_{ij}$=1 \& $q_{ij}$=0, the corresponding BL discharges and if $k_{ij}$=0 \& $q_{ij}$=1, the corresponding BLB discharges. If $k_{ij}$=$q_{ij}$, BL \& BLB maintain their precharge levels proportional to the bit significance factor $j$.             

In \underline{step-3}, to calculate the sum of all the differences, the charge-sum (CSUM) is activated. This results in averaging BL charges on the SLP and BLB charges on SLN through transmission gates at the top. Therefore,
\begin{subequations}
\begin{equation}
    V_{SLP} \approx \frac{1}{n \times p} \sum_{i=0}^{n-1} \sum_{j=0}^{p-1} \big(1-\mathbbm{1}_{k_{ij}=1, q_{ij}=0}\big) \times V_{P,j}
\end{equation}
\begin{equation}
    V_{SLN} \approx \frac{1}{n \times p} \sum_{i=0}^{n-1} \sum_{j=0}^{p-1} \big(1-\mathbbm{1}_{k_{ij}=0, q_{ij}=1}\big) \times V_{P,j}
\end{equation}
\end{subequations}
Here, $\mathbbm{1}$ is the indicator function that is one only when the identity in the subscript is true and zero otherwise. $\times V_{P,j}$ is the column precharge voltage for j$^{th}$ significance bits. Therefore, the average voltage of SLP and SLN, $(V_{SLP} + V_{SLN})/2$, follows the Manhattan distance of $\mathcal{K}$ and $\mathcal{Q}$. In \underline{step-4}, the average voltage is digitized and multiplied with Bayesian optimization learned weights. If a key vector cannot fit in one memory array, it can be partitioned and processed in parallel across several memory arrays. The weighted distance from all arrays is combined, and the minimum distance index is searched by serially scanning all stored key vectors.    

Fig. 6(b) shows the SLP/SLN voltage distribution under the process variability while considering minimum-sized NMOS and PMOS with $\sigma_{VTH}$ = 60 mV) on 32$\times$32 bitcell array. Since the minimum sum line voltage difference ($\Delta$SL) is at least 28 mV, the analog output can be accurately digitized with a 5-bit ADC. Fig. 6(c) shows the distribution of energy among various operations. ADC's energy is estimated from \cite{jiang2021analog}. Peripherals and precharge energy is simulated using HSPICE based on 16 nm-LSTP predictive technology models in \cite{asuptm}. The energy dissipation of digital logic operations in step-4 is estimated based on \cite{10.1155/2015/726175}. One key/query-vector matching operation on 32$\times$32 consumes $\sim$4.7 pJ energy in our 16 nm design.


\begin{table}[t]
    \setlength{\tabcolsep}{4pt}
    \renewcommand{\arraystretch}{1.2}
    \centering
    \caption{\textbf{Table I: MBI \textit{vs.} CIM on MNIST}}
    \begin{tabularx}{\columnwidth}{c|c|c|c|c}
        \hline
        \textbf{Metric} & \textbf{MBI} & \textbf{CIM \cite{9437302}} & \textbf{CIM \cite{8915834}} & \textbf{CIM \cite{8579538}} \\
        \hline
        Technology & 16nm & 28nm & 55nm & 65nm \\
        Model & RAM & MLP & ResNet20 & LeNet \\
        \#Weights & -- & 930,816 & 272,650 & 61,706 \\
        Weight precision & -- & 8-bit & 5-bit & 1-bit \\
        \#MACs & -- & 930,816 & 114,947,584 & 411,240 \\
        TOPS/W & -- & 18.45 & 18.37 & 40.3 \\
        \hline
        Storage (MB) & 6.21 & 0.88 & 0.162 & 0.007 \\
        Accuracy (\%) & 93.04$^*$ & 99.15 & 99.52 & 98.0 \\
        Energy/Inference & 13.16 nJ$^{**}$ & 100.9 nJ & 12.51 $\mu$J & 20.4 nJ \\
        (projected to 16nm)$^{***}$ & -- & (32.94 nJ) & 1.05 $\mu$J & 1.24 nJ \\
        \hline
    \end{tabularx}
    \begin{flushleft}
        \textbf{Comments:} \\
        $^*$The accuracy is extracted under mixed-MBI at a threshold value 5 where 69.65\% data can be processed with MBI. \\
        $^{**}$Energy/Inference in MBI = \# of glimpses (= 5) $\times$ \# of avg. levels in LUT-tree (= 3.5) $\times$ \# of key vectors per LUT (= 32) $\times$ \# of splits of key vectors (= 5) $\times$ avg. energy per comparison (= 4.7 pJ) \\
        $^{***}$ The energy of other works, $E_T$, in technology node $T$ is optimistically projected to 16 nm as $E_T \times (16 \text{nm}/T)^2$.
    \end{flushleft}
\end{table}

\subsection{Compute-in-Memory \textit{vs.} Memorization-based Inference}
Compute-in-memory (CIM) has become a predominant approach to improve the energy efficiency of deep learning by leveraging the same memory structure for storage and computations \cite{nasrin2021mf, shukla2020mc, nasrin2022enos}. Table I compares both paradigms, CIM and MBI, for the MNIST characterization test case, which differ in many key aspects: \textit{Firstly}, unlike CIM, MBI only constrains input bit precision; weights in MBI need not quantize. The underlying RAM architecture can be simulated at full precision to distill LUTs. \textit{Secondly}, MBI is agnostic to necessary multiply-accumulate (MAC) operations, a key metric of various neural network architectures. Specifically, complex models such as ResNet20 in the table can generalize better to more complex tasks but demand many MACs. MBI is more suited for the distillation of such complex models.In our approach, we have applied threshold value to determine the amount of data to be processed by the MBI. When the threshold value is set to be highly stringent, only a small number of images undergo processing by MBI, resulting in nearly perfect accuracy. However, if the threshold value is set to be less stringent, a larger proportion of images are processed by MBI, but the accuracy may slightly decrease. \textit{Thirdly}, MBI has a constant storage overhead that doesn't grow proportionally to predictive model complexity. In Table I, for MNIST, the storage overheads of MBI are significantly worse than CIM; however, for more complex models, MBI can be significantly more storage efficient by avoiding the storage of model parameters. \textit{Fourthly}, by avoiding computations and utilizing only lookups, MBI achieves significantly lower energy per input image inference (Energy/Inference in Table I) even when the other works are optimistically projected to 16 nm (see comments in the table). MBI requires higher energy than one-bit weight CIM design \cite{8579538}; however, the applicability of single-bit weight models is only limited to simple tasks. More complex tasks, such as object localization or ImageNet classification, require sufficiently high precision. 

\section{Conclusions}
We have introduced a novel memory-based inference approach that allows model size-agnostic inference under constant time and storage budget. Our method condenses predictions from a trained model into LUT. During inference, the LUT is searched for the closest matching key vector to a given input glimpse. By performing memorization-based predictions on multiple glimpses, the final prediction is obtained. Compared to competitive compute-in-memory (CIM) approaches, MBI improves energy efficiency by $\sim$2.7$\times$ than multilayer perceptions (MLP)-CIM and by $\sim$83$\times$ than ResNet20-CIM for MNIST character recognition. 

\bibliographystyle{IEEEtran}
\bibliography{main.bib}
\vfill

\end{document}